\documentclass{article}

\usepackage{microtype}
\usepackage{graphicx}
\usepackage{booktabs}
\usepackage{xcolor}
\usepackage{listings}
\usepackage{latexsym}
\usepackage[T1]{fontenc}
\usepackage[utf8]{inputenc}
\usepackage{inconsolata}
\usepackage{amsmath,amssymb,mathtools}
\usepackage{algorithm}
\usepackage{algorithmic}
\usepackage{placeins}
\usepackage{float}
\usepackage{xspace}
\usepackage{url}
\usepackage{hyperref}

\usepackage[preprint]{icml2026}

\usepackage[capitalize,noabbrev]{cleveref}

\icmltitlerunning{Compositional Consistency-Guided Decoding}

\DeclareRobustCommand{\True}{\ifmmode\text{\textsc{True}}\else\textnormal{\textsc{True}}\xspace\fi}
\DeclareRobustCommand{\False}{\ifmmode\text{\textsc{False}}\else\textnormal{\textsc{False}}\xspace\fi}
\DeclareRobustCommand{\Unknown}{\ifmmode\text{\textsc{Unknown}}\else\textnormal{\textsc{Unknown}}\xspace\fi}
\DeclareRobustCommand{\Yes}{\ifmmode\text{\textsc{Yes}}\else\textnormal{\textsc{Yes}}\xspace\fi}
\DeclareRobustCommand{\No}{\ifmmode\text{\textsc{No}}\else\textnormal{\textsc{No}}\xspace\fi}

\newcommand{\NegMap}{\mathsf{NegMap}}
\newcommand{\Entails}{\models}
\newcommand{\NotEntails}{\not\models}

\definecolor{PromptBg}{HTML}{F8F8F4}
\definecolor{PromptRule}{HTML}{D9D6CC}
\lstdefinestyle{promptstyle}{%
  basicstyle=\ttfamily\scriptsize,
  backgroundcolor=\color{PromptBg},
  rulecolor=\color{PromptRule},
  frame=single,
  framerule=0.25pt,
  framesep=3pt,
  xleftmargin=0pt,
  xrightmargin=0pt,
  breaklines=true,
  breakatwhitespace=false,
  columns=fullflexible,
  keepspaces=true,
  showstringspaces=false,
  tabsize=2,
  aboveskip=0.45em,
  belowskip=0.45em
}
\lstnewenvironment{promptblock}{\lstset{style=promptstyle}}{}
\begin{document}

\twocolumn[
\icmltitle{Compositional Consistency-Guided Decoding for Three-Way Logical Question Answering}

\begin{icmlauthorlist}
\icmlauthor{Tianyi Huang}{ryquo}
\icmlauthor{Ming Hou}{appin}
\icmlauthor{Jiaheng Su}{appin}
\icmlauthor{Yutong Zhang}{appin}
\icmlauthor{Ziling Zhang}{appin}
\end{icmlauthorlist}

\icmlaffiliation{ryquo}{Ryquo}
\icmlaffiliation{appin}{App-In Club}

\icmlcorrespondingauthor{Tianyi Huang}{tianyi@ryquo.com}

\icmlkeywords{Compositional reasoning, logical question answering, consistency, abstention, inference-time verification, large language models}

\vskip 0.3in
]

\printAffiliationsAndNotice{}

\begin{abstract}
Logical question answering offers a rare advantage for evaluating LLM reasoning: for some input transformations, the required relationship between outputs is exact.
In three-way logical QA, a consistent premise set assigns a hypothesis $H$ one of \True, \False, or \Unknown, and the label for the negated hypothesis $\neg H$ must follow a fixed negation map.
Yet black-box LLMs can treat these paired queries as unrelated, producing inconsistent labels or retreating to \Unknown even when the premises decide one side.
We turn this built-in structure into a test-time decoding rule.
CGD-PD queries $H$ and $\neg H$, projects candidate labels onto the negation-consistency constraint, and invokes targeted binary entailment probes only when the three-way signal remains ambiguous.
The method is training-free, solver-free, and compatible with ordinary structured prompting.
On FOLIO's first-order logic fields, CGD-PD converts a simple consistency diagnostic into a practical reliability layer: it reduces epistemic abstention while preserving the task's genuine \Unknown option.
The results suggest a broader design principle for LLM reasoning systems: when a task exposes known compositional relations, decoding should use them rather than evaluate each prompt in isolation.
\end{abstract}

\section{Introduction}
Compositionality---the ability to construct and reason about complex objects from reusable parts and their relations---is often treated as a central ingredient for systematic generalization \citep{fodor1988connectionism,lake2018generalization,hupkes2020compositionality}.
For foundation models, a basic test of compositional behavior is whether predictions on related inputs respect known relationships between those inputs.
In logical question answering (QA), such relationships are especially useful because they are not merely stylistic perturbations: they impose hard constraints on the outputs.
This makes logic-oriented QA a useful setting for studying whether a model's answers are accurate, stable, and semantically consistent under simple logical transformations \citep{bowman2015snli,williams2018multinli,ribeiro2020beyond,cho2025metamorphic}.

This paper focuses on three-way logical QA, where a set of assumptions $S$ and a hypothesis $H$ are paired with one of three labels, $\{\True,\False,\Unknown\}$, corresponding to $S\Entails H$, $S\Entails \neg H$, or neither.
Three-way labeling is attractive because it distinguishes contradiction from underspecification.
However, in LLM-based pipelines, \Unknown can play two roles.
It can represent genuine logical underspecification, but it can also act as an implicit abstention when the model is uncertain, inconsistent, or sensitive to phrasing \citep{kadavath2022language,cho2025metamorphic}.
This makes evaluation subtle: a model that predicts \Unknown frequently may appear conservative, while still failing to answer cases whose label is logically determined.

A second issue is more structural.
For consistent premises, the label for $H$ determines the label for $\neg H$ through a fixed negation map. Under this map, \True and \False swap, while \Unknown remains \Unknown.
Thus, three-way logical QA contains a simple compositional consistency relation between two views of the same underlying inference problem.
Standard prompting usually ignores this relation by treating $H$ and $\neg H$ as independent inputs.

\paragraph{From a diagnostic to a decoder.}
The coupling induced by negation creates a constrained two-view inference problem: asking about $H$ and $\neg H$ provides two noisy observations of the same underlying logical state.
If one view is decisive and the other is ambiguous, the decisive view can be mapped back through the negation relation; if both views are ambiguous, a narrower entailment question can test whether either side is supported.
This suggests an inference-time layer that is more structured than repeated sampling of a single prompt, but much lighter than external proof search.

\paragraph{Approach overview.}
We propose \textbf{Consistency-Guided Decoding with Proof-Driven Disambiguation (CGD-PD)}.
Given $(S,H)$, CGD-PD first queries the same 3-way classifier on $H$ and on a mechanically negated form $\neg H$.
It accepts pairs that already satisfy the negation map, applies a targeted \Unknown re-evaluation only to sides that abstain, uses binary entailment probes when both sides remain \Unknown, and adjudicates only when both sides are decisive but contradictory.
The final decision is always interpreted through the negation-consistency constraint.
CGD-PD is training-free, uses no external solver, and can be applied to black-box models that support structured prompting.

\paragraph{Empirical focus.}
We evaluate CGD-PD on FOLIO's first-order logic fields \citep{han2022folio}, where negation can be applied mechanically and the three-way label is well defined.
This controlled setting isolates whether a known compositional relation can improve inference-time reliability when the negation operation is unambiguous.
Across two instruction-tuned API models under the same strict prompt family, CGD-PD improves validation accuracy and reduces epistemic \Unknown.
The diagnostics show that the gains are concentrated on determinate examples that the single-call classifier had treated as \Unknown, while the main remaining tension is protecting genuinely underspecified examples from over-resolution.

\paragraph{Contributions.}
This work makes three contributions:
\begin{itemize}
\item We formulate three-way logical QA as a compositional consistency problem in which predictions for $H$ and $\neg H$ must satisfy a deterministic label relation.
\item We introduce CGD-PD, a lightweight test-time wrapper that combines neural 3-way classification, symbolic negation projection, targeted \Unknown re-evaluation, and binary entailment probes.
\item We provide a focused evaluation on FOLIO's formal fields, including accuracy, abstention, coverage, and gold-\Unknown diagnostics that separate epistemic recovery from overconfident resolution.
\end{itemize}

\section{Problem Setting}
Each instance consists of a premise set $S$ and a hypothesis $H$.
The task is to output one of three labels:
\begin{align*}
\True &: \quad S \Entails H,\\
\False &: \quad S \Entails \neg H,\\
\Unknown &: \quad S \NotEntails H \ \text{and}\ S \NotEntails \neg H.
\end{align*}
We assume $S$ is logically consistent, as is typical for curated benchmarks such as FOLIO \citep{han2022folio}.

\subsection{Negation mapping and consistency}
Negation induces a deterministic mapping on labels:
\[
\begin{aligned}
\NegMap(\True) &= \False,\\
\NegMap(\False) &= \True,\\
\NegMap(\Unknown) &= \Unknown.
\end{aligned}
\]
If a system outputs labels for both $H$ and $\neg H$, negation consistency requires
\begin{equation}
\label{eq:neg-consistency}
y(\neg H)=\NegMap(y(H)).
\end{equation}
Although \eqref{eq:neg-consistency} is elementary, LLMs can violate it when prompted independently.
CGD-PD treats \eqref{eq:neg-consistency} as a hard constraint for its final decision.

\subsection{Epistemic \Unknown versus genuine \Unknown}
On datasets with an explicit \Unknown label, \Unknown can arise for two different reasons:
\begin{itemize}
\item \textbf{Genuine underspecification:} the premises do not entail either side.
\item \textbf{Epistemic \Unknown:} the model returns \Unknown due to uncertainty, instability, or conservative behavior even when one side is entailed.
\end{itemize}
This distinction matters because epistemic \Unknown reduces accuracy and coverage without reflecting the task semantics.
Our goal is to reduce epistemic \Unknown while preserving genuine underspecification whenever the available probes do not support a decisive label.

\section{Method: CGD-PD}
CGD-PD is a test-time wrapper around an LLM that can be prompted as a 3-way classifier.
It uses three reusable components: (i) a base 3-way classifier, (ii) a targeted \Unknown fixer, and (iii) binary entailment probes used as a lightweight proof check.
The final decoder composes these components with the symbolic negation map in \eqref{eq:neg-consistency}.

\subsection{Base 3-way classifier}
We define $\mathsf{Classify}(S,H)\in\{\True,\False,\Unknown\}$ by prompting an LLM to output a structured label.
In our experiments, the prompt requires a strict schema so that the model returns only one of the three labels.
Because \Unknown can act as abstention, we enable a soft bias-away-from-\Unknown instruction in the base prompt.
In the implementation this is controlled by an \texttt{unknown\_penalty} configuration parameter, set to $0.5$ in the reported runs; the scalar is not an API-level logit bias and is not exposed to the model as a numerical weight.
Instead, it adds the qualitative instruction that \Unknown should be used only when the premise block genuinely leaves both sides possible.\footnote{Appendix~\ref{app:ablations} reports development-stage variants that helped motivate the final selective design.}
Appendix~\ref{app:prompts} provides the prompt templates for the base classifier, the \Unknown fixer, the binary entailment probe, and the adjudicator.

\subsection{Composing hypothesis and negation probes}
Given $(S,H)$, CGD-PD first queries the base classifier on $H$ and on a mechanically negated hypothesis $\neg H$.
In practice, we represent negation with a canonical wrapper (e.g., \texttt{NOT:} $H$) and explicitly define its semantics in the prompt.
We denote the two outputs by $y_H=\mathsf{Classify}(S,H)$ and $y_{\neg H}=\mathsf{Classify}(S,\neg H)$.
If $y_{\neg H}=\NegMap(y_H)$ and at least one side is decisive, we accept the pair as negation-consistent and return $y_H$.
Otherwise, we proceed to disambiguation or adjudication.

\subsection{Targeted \Unknown fixing}
When one side returns \Unknown, we do not immediately force a decision.
Instead, we run a targeted prompt $\mathsf{FixUnknown}(S,H)$ that asks the model to re-evaluate the same premise block and hypothesis under the original 3-way decision rule.
The fixer treats \Unknown as a last resort: it should return \True or \False when the premise block contains a decisive logical path, and return \Unknown only when a specific required fact is missing.
When it keeps \Unknown, the prompt asks it to name the missing premise and, when possible, quote the part of the premise block that indicates the gap.
When it returns a decisive label, the auxiliary fields are set to null.

After applying the fixer on the \Unknown side(s), we first check whether the updated pair is already negation-consistent.
If so, we return the label for $H$.
If exactly one side is decisive and the other remains \Unknown, we project the remaining side through \eqref{eq:neg-consistency} and return the corresponding label for $H$.
This projection is rule-constrained but not a formal guarantee: it depends on the reliability of the decisive model call that triggered it.

\subsection{Proof-driven disambiguation via binary entailment probes}
If both sides remain \Unknown after targeted fixing, we use binary entailment probes $b_H=\mathsf{EntailsYesNo}(S,H)\in\{\Yes,\No\}$ and $b_{\neg H}=\mathsf{EntailsYesNo}(S,\neg H)\in\{\Yes,\No\}$.
The binary probe asks a narrower question than 3-way classification and removes \Unknown as an output option.
This can reveal cases where the 3-way prompt used \Unknown as a default, but it can also overcommit; the decoder therefore uses the two binary answers only when they agree with a consistent one-sided entailment pattern.
Specifically, if $(b_H,b_{\neg H})=(\Yes,\No)$, CGD-PD returns \True; if $(b_H,b_{\neg H})=(\No,\Yes)$, it returns \False; otherwise, it keeps \Unknown.
This includes the conflict case where both probes say \Yes; rather than privileging one side arbitrarily, the method abstains.

We call this step proof-driven because focused entailment questions provide lightweight evidence that one side is derivable from $S$.
It is not a formal proof system, and we do not claim solver-level guarantees.

\subsection{Adjudication for inconsistent decisive pairs}
Finally, if $y_H$ and $y_{\neg H}$ are both decisive but violate \eqref{eq:neg-consistency}, we invoke a lightweight adjudicator prompt. It chooses between the two consistent assignments, returning either $y_H$ or $\NegMap(y_{\neg H})$.
The adjudicator is used only when the classifier has produced a contradictory decisive pair.

\subsection{Algorithm and compute}
Algorithm~\ref{alg:cgd-pd} summarizes CGD-PD.
The wrapper makes two calls in the common case and up to six calls when both fixers and both binary probes are required.
In our FOLIO validation runs, it uses 4--5 calls on average (Section~\ref{sec:results}).

\begin{algorithm}[!t]
\caption{CGD-PD: Consistency-Guided Decoding with Proof-Driven Disambiguation}
\label{alg:cgd-pd}
\small
\begin{algorithmic}[1]
\REQUIRE Premises $S$, hypothesis $H$, negation mapping $\NegMap$
\STATE $y_H \leftarrow \mathsf{Classify}(S,H)$
\STATE $y_{\neg H} \leftarrow \mathsf{Classify}(S,\neg H)$ \hfill (mechanical negation)
\IF{$y_{\neg H} = \NegMap(y_H) \land \neg (y_H=\Unknown \land y_{\neg H}=\Unknown)$}
  \STATE \textbf{return} $y_H$
\ENDIF
\IF{$y_H=\Unknown$}
  \STATE $y_H \leftarrow \mathsf{FixUnknown}(S,H)$
\ENDIF
\IF{$y_{\neg H}=\Unknown$}
  \STATE $y_{\neg H} \leftarrow \mathsf{FixUnknown}(S,\neg H)$
\ENDIF
\IF{$y_H\neq \Unknown \land y_{\neg H}\neq \Unknown \land y_{\neg H}=\NegMap(y_H)$}
  \STATE \textbf{return} $y_H$ \hfill (consistent after fixing)
\ENDIF
\IF{$y_H\neq \Unknown \land y_{\neg H}=\Unknown$}
  \STATE \textbf{return} $y_H$ \hfill (project via $\NegMap$)
\ELSIF{$y_H=\Unknown \land y_{\neg H}\neq \Unknown$}
  \STATE \textbf{return} $\NegMap(y_{\neg H})$
\ENDIF
\IF{$y_H=\Unknown \land y_{\neg H}=\Unknown$}
  \STATE $b_H \leftarrow \mathsf{EntailsYesNo}(S,H)$
  \STATE $b_{\neg H} \leftarrow \mathsf{EntailsYesNo}(S,\neg H)$
  \IF{$b_H=\Yes \land b_{\neg H}=\No$}
    \STATE \textbf{return} \True
  \ELSIF{$b_H=\No \land b_{\neg H}=\Yes$}
    \STATE \textbf{return} \False
  \ELSE
    \STATE \textbf{return} \Unknown
  \ENDIF
\ENDIF
\STATE \textbf{return} $\mathsf{Adjudicate}(S,H,y_H,y_{\neg H})$ \hfill (decisive conflict)
\end{algorithmic}
\end{algorithm}

\section{Experimental Setup}
\label{sec:setup}
Our experiments ask whether a minimal inference-time layer can improve task accuracy and reliability diagnostics. We focus on consistency under negation and epistemic abstention.

\subsection{Dataset: FOLIO}
We evaluate on FOLIO \citep{han2022folio}, which pairs short natural-language stories with first-order logic annotations for both the premises and the hypothesis.
FOLIO labels each hypothesis as one of \True, \False, or \Unknown under standard first-order semantics.
The FOL fields reduce ambiguity in negation, allowing us to mechanically represent $\neg H$ in a controlled way.

\paragraph{Input representation.}
Unless otherwise stated, we run CGD-PD on the FOL fields: premises are the dataset-provided FOL formulas, and hypotheses are the dataset-provided FOL conclusions.
This choice makes the negation operation unambiguous and allows us to isolate the consistency mechanism.
It also defines the scope of the evaluation: natural-language negation scope is an important and harder setting, and robustness there should be tested directly when negation must be generated or interpreted from text.

\subsection{Models and decoding}
We evaluate two instruction-tuned LLMs accessed via API: \texttt{gpt-5.2} and \texttt{claude-sonnet-4-5}.
All prompts require structured outputs with an explicit schema, and temperature is set to 0 to reduce sampling variability.
We do not include chain-of-thought in the main protocol, which reduces format variability and keeps the prompt family consistent across methods.
This protocol is intentionally conservative and should not be interpreted as measuring the best achievable performance of either model on FOLIO; in particular, a more permissive prompt or explicit reasoning format could change the single-call baseline and the relative gain from CGD-PD.

\subsection{Baselines and development-stage variants}
Our primary comparison is against \textbf{Single}, which makes one call to $\mathsf{Classify}(S,H)$.
Appendix~\ref{app:ablations} also reports two development-stage variants used during method design:
\textbf{gate3}, a 3-call gating-style variant without binary entailment probes, and \textbf{CGR3}, a 3-call consistency-guided redundancy variant without the proof-driven disambiguation stage.
CGD-PD is not intended to replace ensembling methods such as self-consistency \citep{wang2022selfconsistency}; it addresses a complementary axis, namely logical coupling across related probes.
Matched-call self-consistency is a natural complementary baseline for call-budget efficiency; accordingly, our claims focus on the reliability effects of using logically coupled probes rather than on optimal call allocation.

\subsection{Metrics}
We report:
\begin{itemize}
\item \textbf{Accuracy}: fraction of hypotheses whose predicted label matches the gold label.
\item \textbf{\Unknown rate}: fraction of predictions that are \Unknown.
\item \textbf{Epistemic \Unknown rate}: fraction of gold determinate examples (\True or \False) predicted as \Unknown.
\item \textbf{Compute}: average number of model calls per example.
\end{itemize}
We report 95\% confidence intervals for key deltas using paired bootstrap resampling over examples.

\section{Results}
\label{sec:results}

\subsection{Main quantitative results}
Table~\ref{tab:main} summarizes validation performance on FOLIO's FOL fields (204 examples).
Using the same structured prompt family for the single-call and wrapped systems, CGD-PD improves accuracy on both models while reducing \Unknown predictions.

\begin{table*}[t]
\caption{FOLIO validation results on FOL fields (204 validation examples). Accuracy, \Unknown rate, and epistemic \Unknown rate are percentages; epistemic \Unknown is the fraction of gold \True/\False examples predicted as \Unknown. ``Calls'' is the mean number of wrapper model invocations per example.}
\label{tab:main}
\centering
\small
\setlength{\tabcolsep}{5pt}
\begin{tabular}{llcccc}
\toprule
Model & Method & Acc. (\%) & \Unknown (\%) & Epist.\ \Unknown (\%) & Calls \\
\midrule
GPT-5.2 & Single & 63.7 & 57.4 & 41.5 & 1.00 \\
GPT-5.2 & CGD-PD & \textbf{68.1} & \textbf{53.9} & \textbf{36.3} & 4.36 \\
\midrule
Claude Sonnet 4.5 & Single & 42.2 & 75.5 & 72.6 & 1.00 \\
Claude Sonnet 4.5 & CGD-PD & \textbf{49.0} & \textbf{58.8} & \textbf{53.3} & 4.91 \\
\bottomrule
\end{tabular}
\end{table*}

\paragraph{Confidence intervals.}
The paired bootstrap confidence intervals exclude zero for both models.
For GPT-5.2, CGD-PD yields a +4.4 point accuracy gain (95\% CI: +1.5 to +7.4) and a $-3.4$ point reduction in \Unknown rate (95\% CI: $-6.4$ to $-0.5$).
For Claude Sonnet~4.5, accuracy improves by +6.8 points (95\% CI: +3.4 to +10.3) and \Unknown rate drops by $-16.7$ points (95\% CI: $-21.6$ to $-11.8$).
The Claude single-call baseline is notably conservative under this strict no-chain-of-thought schema, with 75.5\% \Unknown predictions.
We therefore interpret the Claude result as evidence that CGD-PD can recover from epistemic abstention under this prompt family, not as a claim about the model's best possible FOLIO performance.

\section{Analysis}
\label{sec:analysis}
We analyze how CGD-PD changes predictions beyond the aggregate accuracy numbers.
Because the method reduces abstention, we separately examine final accuracy, coverage, and the reliability of non-\Unknown decisions.
We use ``answered accuracy'' to mean accuracy conditional on a non-\Unknown prediction; this is not probability calibration, but it tests whether the wrapper merely converts \Unknown into decisive labels or also preserves reliability among the labels it chooses to answer.

\subsection{Confusion matrices: resolving epistemic \Unknown}
Figure~\ref{fig:confusion} in Appendix~\ref{app:confusion} shows row-normalized confusion matrices for GPT-5.2 and Claude Sonnet~4.5.
Two patterns stand out:
\begin{itemize}
\item \textbf{Reduced abstention on determinate cases.} For both models, many gold \True and \False examples are predicted as \Unknown by the single classifier. CGD-PD moves some of these to the correct decisive label, reducing epistemic \Unknown.
\item \textbf{Boundary between recovery and over-resolution.} For GPT-5.2, the fraction of gold \Unknown predicted as \Unknown is unchanged. For Claude Sonnet~4.5, CGD-PD predicts \Unknown less often even when the gold label is \Unknown, introducing additional \True/\False errors. The net effect remains positive because the recovered determinate cases dominate, but this diagnostic identifies where a future variant should be more selective.
\end{itemize}

\subsection{Coverage and decisive-label reliability}
Table~\ref{tab:coverage} summarizes coverage, answered accuracy, and errors on gold-\Unknown examples.
For GPT-5.2, CGD-PD increases coverage from 42.6\% to 46.1\% and answered accuracy from 79.3\% to 83.0\%.
For Claude Sonnet~4.5, coverage increases more substantially, from 24.5\% to 41.2\%, while answered accuracy increases from 60.0\% to 61.9\%.
These numbers suggest that CGD-PD is not simply replacing abstentions with lower-quality decisive labels in this setting.
The gold-\Unknown columns also identify the key selectivity challenge: for Claude, the number of genuinely underspecified examples converted to a decisive label rises from 13 to 21.
The final-output error profile is mildly asymmetric for Claude (13 \True errors vs. 8 \False errors on gold \Unknown), making probe-level label bias a useful target for follow-up analysis.

\begin{table*}[t]
\caption{Coverage and decisive-label reliability on FOLIO validation (204 examples). Coverage, answered accuracy, and Gold U kept are percentages; Gold U$\to$T/F are counts of genuinely underspecified examples predicted as decisive labels. T/F/U denote \True, \False, and \Unknown.}
\label{tab:coverage}
\centering
\small
\setlength{\tabcolsep}{5pt}
\begin{tabular}{llccccc}
\toprule
Model & Method & Cov. & Ans. acc. & Gold U kept & Gold U$\to$T & Gold U$\to$F \\
\midrule
GPT-5.2 & Single & 42.6 & 79.3 & 88.4 & 3 & 5 \\
GPT-5.2 & CGD-PD & 46.1 & 83.0 & 88.4 & 4 & 4 \\
\midrule
Claude Sonnet 4.5 & Single & 24.5 & 60.0 & 81.2 & 9 & 4 \\
Claude Sonnet 4.5 & CGD-PD & 41.2 & 61.9 & 69.6 & 13 & 8 \\
\bottomrule
\end{tabular}
\end{table*}

\subsection{How often does proof-driven disambiguation trigger?}
CGD-PD is designed to spend extra calls primarily when the model abstains or is inconsistent.
On FOLIO validation, CGD-PD uses the maximum six calls on 54\% of GPT-5.2 examples and 61\% of Claude examples, reflecting the prevalence of \Unknown outputs in this task.
On GPT-5.2, CGD-PD changes the single-model prediction on 15/204 examples; on Claude, it changes 34/204, largely by converting \Unknown into decisive outputs.
These aggregate change counts do not isolate which internal branch caused each change, so they should be read together with the confusion matrices rather than as an adjudicator-specific audit.

\subsection{What the current robustness evidence establishes}
The paired bootstrap intervals and coverage diagnostics provide evidence that the observed gains are stable under resampling of this validation split and are not merely due to converting \Unknown into low-quality decisive answers.
The evidence is strongest for the controlled setting studied here: FOL-formula inputs, mechanical negation, two API models, and one strict structured-output prompt family.
The most informative next checks are therefore clear: allocate the same call budget to self-consistency or related inference-time baselines, and test natural-language FOLIO inputs where negation scope must be interpreted rather than mechanically specified.

\section{Related Work}
\label{sec:related}

\paragraph{Compositionality and modular reasoning.}
Compositional generalization has long been studied as a challenge for neural systems \citep{fodor1988connectionism,lake2018generalization,hupkes2020compositionality,keysers2020measuring}.
A related line of work studies modular architectures that assemble reusable components for complex tasks \citep{andreas2016neural}.
CGD-PD is much narrower than these broad programs: it does not learn compositional representations or modules.
Instead, it uses a fixed logical relation, negation, to compose multiple inference-time views into a more consistent final decision.

\paragraph{Logic-oriented QA benchmarks.}
Standard NLI benchmarks such as SNLI and MultiNLI \citep{bowman2015snli,williams2018multinli} have been central to evaluating entailment, but natural-language datasets can conflate linguistic heuristics with logical competence.
Logic-focused resources add structure to reduce ambiguity, including RuleTaker and ProofWriter \citep{clark2020ruletaker,tafjord2020proofwriter} and more recent benchmarks targeting formal reasoning.
FOLIO \citep{han2022folio} is directly aligned with our setting because it provides first-order logic annotations for both premises and hypotheses together with explicit three-way labels, enabling controlled negation-based transformations.
Related resources such as P-FOLIO \citep{han2024pfolio} and LogicBench \citep{parmar2024logicbench} extend evaluation in complementary directions, including human-written reasoning chains and broader logical phenomena.

\paragraph{Consistency and invariance testing.}
Behavioral testing work such as CheckList \citep{ribeiro2020beyond} argues that accuracy alone can hide brittle behaviors under perturbations.
Metamorphic testing extends this idea by specifying transformations and expected relations between outputs \citep{cho2025metamorphic}.
Our work shares the diagnostic motivation, but uses a small set of logic-specific metamorphic relations to guide inference-time decisions rather than only to evaluate failures.

\paragraph{Inference-time verification.}
Self-consistency and related sampling methods improve reasoning by aggregating repeated generations from the same prompt \citep{wang2022selfconsistency}.
Tree-of-Thoughts and related search-based approaches use larger inference-time budgets to branch and evaluate candidate steps \citep{yao2023tree}.
CGD-PD is complementary: it uses a small number of calls on logically linked prompts and applies a deterministic constraint to project onto a coherent assignment.
The proof-driven disambiguation step is also related to generate-then-verify paradigms and inference-time intervention \citep{li2023inference}.

\paragraph{Constrained decoding, abstention, and structured inference.}
Constrained decoding has a long history in sequence generation, including lexically constrained decoding \citep{hokamp2017lexically,post2018fast} and incremental parsing constraints for text-to-SQL \citep{scholak2021picard}.
These methods primarily enforce syntactic or formal-language constraints.
CGD-PD can be viewed as a lightweight structured inference layer for one semantic constraint.
The explicit \Unknown label also resembles selective prediction and abstention, where models can defer on uncertain inputs \citep{geifman2017selective,kadavath2022language}.

\section{Scope and Limitations}
CGD-PD is a structured inference layer, not a replacement for formal proof search.
Its hard guarantee is that the returned label for $H$ corresponds to a negation-consistent assignment over $(H,\neg H)$; label correctness still depends on the model judgments used by the wrapper and on the implementation of the mechanical negation convention.
The current study isolates this mechanism on one FOLIO validation split, FOL-formula inputs, two API models, and a strict no-chain-of-thought prompt family.
This controlled scope makes the mechanism easier to interpret, while leaving natural-language negation, additional benchmarks, and stronger prompting protocols as important next tests.
CGD-PD also spends additional inference-time calls relative to a single classifier, so matched-budget comparisons with self-consistency and related inference-time baselines would sharpen efficiency claims.
Finally, the present diagnostics are final-output diagnostics: coverage, answered accuracy, and gold-\Unknown error profiles characterize the predictions, but branch-level logs such as fixer changes, adjudicator overrides, and probe-level label biases would give a more detailed account of the mechanism.

\section{Conclusion}
We presented CGD-PD, a training-free inference-time wrapper for three-way logical QA that makes the negation relation explicit at decoding time.
The central idea is simple: predictions for a hypothesis and its negation should not be treated as unrelated samples, because the task already specifies how their labels compose.
On FOLIO's FOL fields, applying this constraint with targeted entailment probes improves both evaluated LLMs and reduces epistemic \Unknown under a strict structured-output protocol.
The analysis clarifies where the method helps---primarily by recovering determinate examples that were previously abstained on---and where care is still needed, especially around genuinely underspecified cases.
More broadly, the work argues for a modest but practical form of reasoning reliability: before adding larger verifiers or new training, inference can often be strengthened by making the task's own compositional structure operational.

\section*{Impact Statement}
By studying how logically grounded consistency checks can expose and reduce specific failure modes in LLM reasoning, this work contributes to efforts toward more reliable and interpretable AI evaluation.
Such methods may be useful in educational tools, analysis assistants, and verification-oriented systems where transparent abstention behavior matters.
Consistency is not the same as truth, however: a system can satisfy a local logical relation and still be wrong about the underlying task.
Deployment should therefore pair inference-time consistency layers with domain-appropriate safeguards and, when possible, independent verification.
Extending these ideas beyond negation to richer logical transformations may support broader settings where abstention, consistency, and reproducibility are central concerns.

\bibliographystyle{icml2026}
\bibliography{main}

\FloatBarrier
\clearpage
\onecolumn
\appendix

\section{Prompt Templates}
\label{app:prompts}

We use structured prompts and parse only the requested JSON fields. The templates below mirror the implementation used for the reported runs, with instance-specific content replaced by placeholders. In the main experiments, \texttt{STORY} contains the FOL premise block; a hypothesis prefixed with \texttt{NOT:} is treated as the mechanical negation of the remaining text. We do not request or parse chain-of-thought.

\subsection{Common Classifier Rules}

The base classifier, \Unknown fixer, and consistency adjudicator prepend the following shared instruction block before their task-specific instructions.

\begin{promptblock}
You are a careful, literal logician.
Task: 3-way logical classification of whether STORY entails HYPOTHESIS.

Valid labels: True, False, Unknown.

Decision rule:
1) True = STORY logically entails HYPOTHESIS.
2) False = STORY logically entails NOT(HYPOTHESIS).
3) Unknown = STORY entails neither HYPOTHESIS nor NOT(HYPOTHESIS).

Constraints:
- Use ONLY the STORY. Do not use outside knowledge or assumptions.
- Apply the decision rule literally.
- Do not use Unknown as a response to uncertainty about your own reasoning.
- Use Unknown only when the STORY genuinely leaves both HYPOTHESIS and NOT(HYPOTHESIS) unentailed.

Negation convention:
- If the HYPOTHESIS begins with 'NOT: ', treat it as the logical negation of the remainder.
\end{promptblock}

\subsection{Input Block}

The implementation appends the premise and hypothesis block below to each prompt.

\begin{promptblock}
STORY:
{STORY}

HYPOTHESIS:
{HYPOTHESIS}
\end{promptblock}

\subsection{Base 3-way Classifier}

\begin{promptblock}
[[TASK:single]]
<COMMON CLASSIFIER RULES>

Output format: Return ONLY a JSON object: {"label": <label>}
Do not include any extra keys, markdown, or commentary.
\end{promptblock}

\subsection{Targeted \Unknown Re-evaluation}

\begin{promptblock}
[[TASK:fix_unknown]]
<COMMON CLASSIFIER RULES>

This prompt is used only after an earlier call returned Unknown.
Re-evaluate the same STORY and HYPOTHESIS under the decision rule above.

If the STORY entails HYPOTHESIS, return True.
If the STORY entails NOT(HYPOTHESIS), return False.
If neither is entailed, return Unknown.

If you return Unknown, provide:
- missing_premise: the specific fact that would be needed to decide the label,
- quote: an exact quote from the STORY showing the relevant gap, or null if no such quote exists.

If you return True or False, set missing_premise=null and quote=null.

Earlier model output:
- prior_label = {PRIOR_LABEL}

Output format: Return ONLY JSON with keys label, missing_premise, quote.
No extra keys, markdown, or commentary.
\end{promptblock}

\subsection{Binary Entailment Probe}

\begin{promptblock}
[[TASK:entails_yesno]]
You are a careful, literal reasoner.
Return ONLY a JSON object with key 'answer' and value 'YES' or 'NO'.

Rules:
- Use ONLY the STORY. Do not use outside knowledge or assumptions.
- Answer YES only if the STORY logically entails the HYPOTHESIS.
- Otherwise answer NO.

STORY:
{STORY}

HYPOTHESIS:
{HYPOTHESIS}

Question: Does the STORY logically entail the HYPOTHESIS?
\end{promptblock}

\subsection{Consistency Adjudicator}

\begin{promptblock}
[[TASK:cgr3_adjudicator]]
<COMMON CLASSIFIER RULES>

Consistency requirement across H and NOT:H:
- If the final label for H is True, then the corresponding label for NOT:H is False.
- If the final label for H is False, then the corresponding label for NOT:H is True.
- If the final label for H is Unknown, then the corresponding label for NOT:H is Unknown.

Earlier one-shot labels:
- label(H) = {Y_H}
- label(NOT:H) = {Y_NOT_H}

The earlier labels may be inconsistent. Use them only as auxiliary signals.
Your job is to output one final label for H that follows the STORY and satisfies the decision rule.

Output format: Return ONLY JSON: {"label": <label>}.
No extra keys, markdown, or commentary.
\end{promptblock}

\section{Confusion Matrices}
\label{app:confusion}
\begin{figure}[H]
\centering
\includegraphics[width=0.6\textwidth]{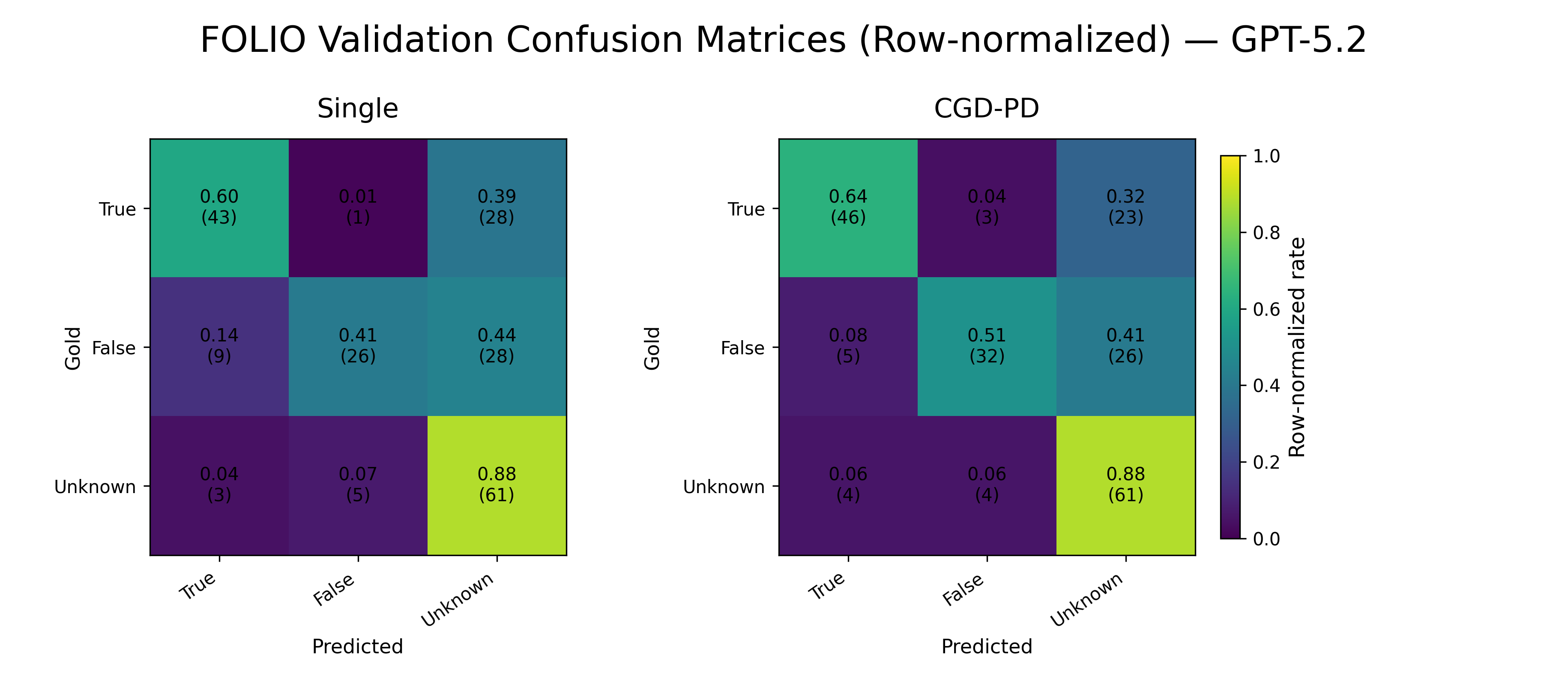}
\vspace{0.5em}
\includegraphics[width=0.6\textwidth]{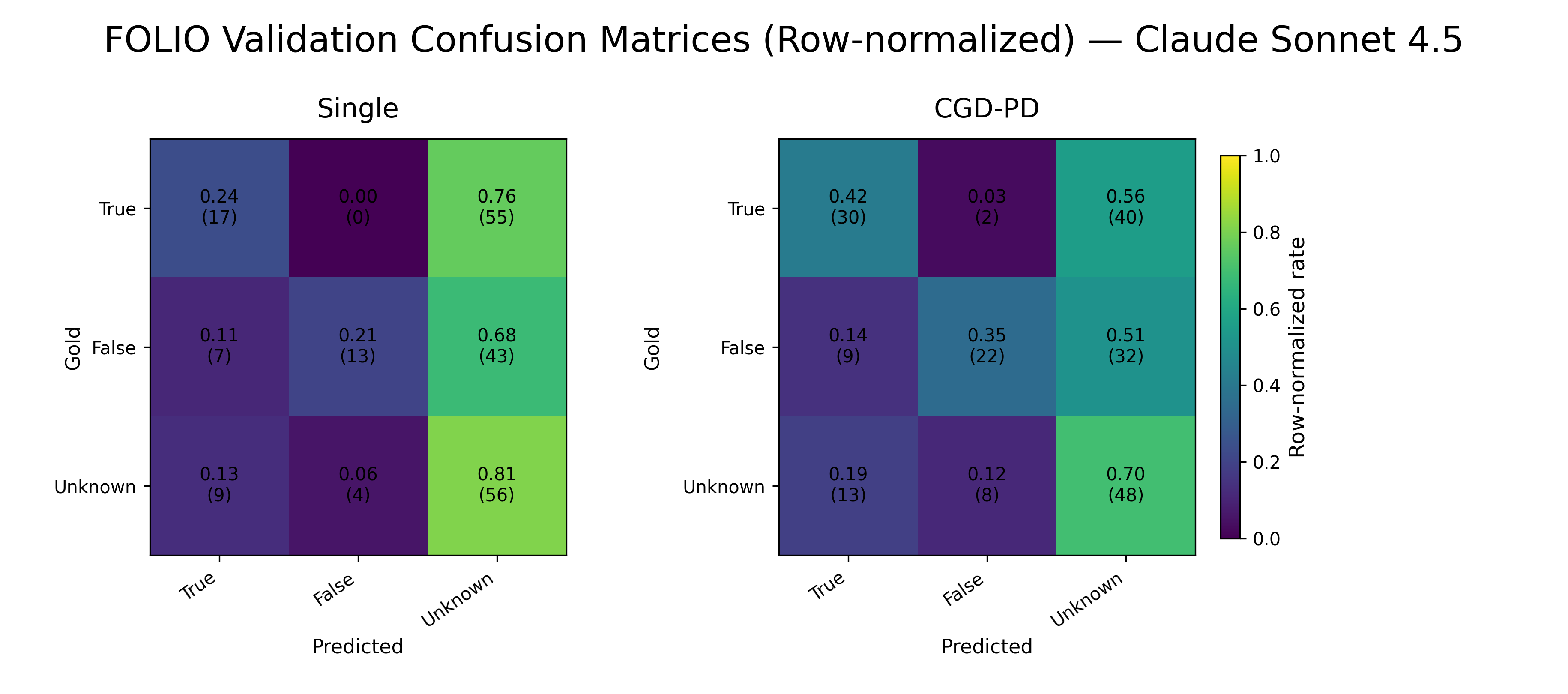}
\caption{Row-normalized confusion matrices on the FOLIO validation split. For each model, the left heatmap is the single-call classifier and the right heatmap is CGD-PD; rows are gold labels and columns are predicted labels. Each cell shows the row-normalized rate with the raw count in parentheses.}
\label{fig:confusion}
\end{figure}

Figure~\ref{fig:confusion} shows row-normalized confusion matrices for the validation split, highlighting how CGD-PD shifts mass away from single-shot \Unknown predictions and, in some cases, corrects overconfident \True/\False errors.

\FloatBarrier

\section{Ablations and Negative Results}
\label{app:ablations}
This appendix summarizes development-stage variants and negative results that informed the final design.
For transparency, Table~\ref{tab:ablations} reports train and validation results, and Table~\ref{tab:qualitative} shows representative validation cases.

\paragraph{What are gate3 and CGR3?}
Both variants were intermediate designs explored in development.
\textbf{gate3} is a 3-call gating baseline that spreads compute across related prompts and applies a lightweight resolution step, but omits CGD-PD's proof-driven binary entailment probes.
\textbf{CGR3} is a 3-call consistency-guided redundancy variant that probes logically related forms and applies a lightweight consistency-aware resolution rule, but omits proof-driven disambiguation.
These variants help isolate what is gained by the final CGD-PD design.

\begin{table}[H]
\caption{Representative GPT-5.2 validation cases illustrating two mechanisms of CGD-PD. The first case preserves a genuine \Unknown when neither side is entailed; the second uses the decisive negated probe to correct the final label for $H$. \textsc{T}/\textsc{F}/\textsc{U} denote \True/\False/\Unknown.}
\label{tab:qualitative}
\centering
\footnotesize
\setlength{\tabcolsep}{4pt}
\begin{tabular}{p{0.12\textwidth} p{0.15\textwidth} c c c p{0.40\textwidth}}
\toprule
Premise excerpt & Hypothesis $H$ & Gold & Single & Ours & Signals used by our method \\
\midrule
\emph{The Picuris Mountains are a mountain range in New Mexico or Texas.} &
The Picuris Mountains are in Texas. &
\textsc{U} &
\textsc{F} &
\textsc{U} &
Probe labels: $y(H)=\Unknown$ and $y(\neg H)=\Unknown$. CGD-PD then runs proof-driven entailment checks and finds $\mathsf{EntailsYesNo}(S,H)=\No$ and $\mathsf{EntailsYesNo}(S,\neg H)=\No$, so it preserves \Unknown. This illustrates that the wrapper need not force a decisive answer when neither side is supported. \\[4pt]

\emph{The Picuris Mountains are a mountain range in New Mexico or Texas.} &
The Picuris Mountains are in New Mexico. &
\textsc{F} &
\textsc{T} &
\textsc{F} &
Probe labels: $y(H)=\Unknown$ and $y(\neg H)=\True$. Because the negated form is decisive, CGD-PD maps that decision back through the negation constraint and returns \False. This illustrates how asymmetric evidence from logically linked probes can correct a single-prompt error. \\
\bottomrule
\end{tabular}
\end{table}

\begin{table}[H]
\caption{Ablations and negative results. Acc = accuracy; Unk = \Unknown rate; Calls/ex = average model calls per example. Validation contains 204 examples in FOLIO v0.0, so \texttt{--max\_examples} is an upper bound rather than an exact size control.}
\label{tab:ablations}
\centering
\small
\setlength{\tabcolsep}{5pt}
\begin{tabular}{llccc ccc}
\toprule
& & \multicolumn{3}{c}{Train (500)} & \multicolumn{3}{c}{Validation (204)}\\
\cmidrule(lr){3-5}\cmidrule(lr){6-8}
Model & Method & Acc & Unk & Calls/ex & Acc & Unk & Calls/ex \\
\midrule
GPT-5.2 & Single & 0.632 & 0.544 & 1.00 & 0.637 & 0.574 & 1.00 \\
GPT-5.2 & gate3 & 0.632 & 0.568 & 3.00 & -- & -- & -- \\
GPT-5.2 & CGR3 (3 calls) & 0.654 & 0.454 & 3.00 & -- & -- & -- \\
GPT-5.2 & \textbf{CGD-PD (final)} & \textbf{0.682} & \textbf{0.396} & 3.85 & \textbf{0.681} & \textbf{0.539} & 4.36 \\
\midrule
Claude Sonnet 4.5 & Single & 0.436 & 0.744 & 1.00 & 0.422 & 0.755 & 1.00 \\
Claude Sonnet 4.5 & \textbf{CGD-PD} & \textbf{0.490} & \textbf{0.546} & 4.80 & \textbf{0.490} & \textbf{0.588} & 4.91 \\
\bottomrule
\end{tabular}
\end{table}

\paragraph{A simple 3-call gate did not reliably help.}
The initial gate-style variant (gate3) matched single-shot accuracy on GPT-5.2 (0.632) while increasing \Unknown (0.544 $\rightarrow$ 0.568) at three times the call budget.
This suggests that spreading compute across probes is insufficient; the integration rule matters.

\paragraph{Redundancy helps only when decoding respects logical structure.}
A three-call variant (CGR3) improved GPT-5.2 train accuracy (0.632 $\rightarrow$ 0.654) and reduced \Unknown (0.544 $\rightarrow$ 0.454), but the gains were sensitive to the resolution rule.
This motivated the final CGD-PD decoder, which exploits asymmetries between $H$ and $\neg H$ while remaining conservative when neither side is supported.

\paragraph{Global pressure against \Unknown is brittle.}
Across early iterations, penalizing \Unknown in the prompt did not consistently reduce \Unknown and sometimes introduced errors on gold-\Unknown examples.
CGD-PD instead applies additional pressure selectively, only when the probe pair provides evidence favoring one side.

\end{document}